\documentclass[10pt,twocolumn,letterpaper]{article}

\usepackage{cvpr}              

\usepackage{graphicx}
\usepackage{amsmath}
\usepackage{amssymb}
\usepackage{booktabs}
\usepackage{subcaption}
\usepackage{multirow}
\usepackage[pagebackref,breaklinks,colorlinks]{hyperref}

\usepackage[capitalize]{cleveref}
\crefname{section}{Sec.}{Secs.}
\Crefname{section}{Section}{Sections}
\Crefname{table}{Table}{Tables}
\crefname{table}{Tab.}{Tabs.}

\def\cvprPaperID{*****} 
\def\confName{CVPR}
\def\confYear{2026}

\begin{document}

\title{SimuFreeMark: A Noise-Simulation-Free Robust Watermarking Against Image Editing}

\author{Yichao Tang, Mingyang Li, Di Miao, Sheng Li, Zhenxing Qian, Xinpeng Zhang$^*$\\
College of Computer Science and Artificial Intelligence, Fudan University\\
{\tt\small yichao\_tang@fudan.edu.cn}
}
\maketitle

\begin{abstract}
The advancement of artificial intelligence generated content (AIGC) has created a pressing need for robust image watermarking that can withstand both conventional signal processing and novel semantic editing attacks. Current deep learning-based methods rely on training with hand-crafted noise simulation layers, which inherently limit their generalization to unforeseen distortions. In this work, we propose \textbf{SimuFreeMark}, a noise-\underline{simu}lation-\underline{free} water\underline{mark}ing framework that circumvents this limitation by exploiting the inherent stability of image low-frequency components. We first systematically establish that low-frequency components exhibit significant robustness against a wide range of attacks. Building on this foundation, SimuFreeMark embeds watermarks directly into the deep feature space of the low-frequency components, leveraging a pre-trained variational autoencoder (VAE) to bind the watermark with structurally stable image representations. This design completely eliminates the need for noise simulation during training. Extensive experiments demonstrate that SimuFreeMark outperforms state-of-the-art methods across a wide range of conventional and semantic attacks, while maintaining superior visual quality. 
\end{abstract}


\section{Introduction}
\label{sec:intro}
The advancement of artificial intelligence generated content (AIGC) has dramatically simplified high-quality image editing, allowing tasks ranging from local content modification to global style transfer with simple instructions~\cite{rombachHighresolutionImageSynthesis2022, zhangArtBankArtisticStyle2024}. However, this productivity gain poses severe challenges to copyright protection, as malicious users can easily alter image content via object replacement or inpainting, potentially removing or distorting embedded watermarks. Traditional watermarking techniques~\cite{gaoEfficientRobustReversible2024, tangNovelRobustReversible2025,tangRobustReversibleWatermarking2024}, designed to withstand pixel-level distortions such as JPEG compression and Gaussian noise, often fail against AIGC-driven semantic manipulations~\cite{huRobustWideRobustWatermarking2025}. Deep learning-based methods~\cite{fangDERODiffusionmodelerasureRobust2024, alamSpecGuardSpectralProjectionbased2025, huoDRSWDualstageRobust2025} show improved robustness against complex physical attacks such as camera capture~\cite{maRoPaSSRobustWatermarking2025, raoDynMarkRobustWatermarking2025, wuSimtorealUnsupervisedNoise2025} and print-scanning~\cite{qinPrintcameraResistantImage2024}. Yet, they still struggle to achieve a satisfactory balance between resisting conventional signal processing attacks and withstanding AIGC semantic edits. This highlights the need for watermarking schemes that can cope with various image distortions.

\begin{figure}[!t]
    \centering
    \setlength{\belowcaptionskip}{0pt}
    \begin{subfigure}{0.48\textwidth}
        \centering
        \includegraphics[width=\textwidth]{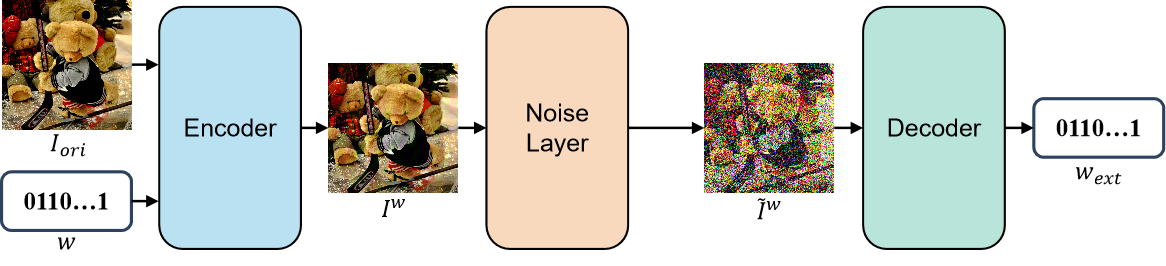}
        \caption{The architecture of END-based framework.}
        \label{Fig:endbased}
    \end{subfigure}
    \hfill
    \begin{subfigure}{0.48\textwidth}
        \centering
        \includegraphics[width=\textwidth]{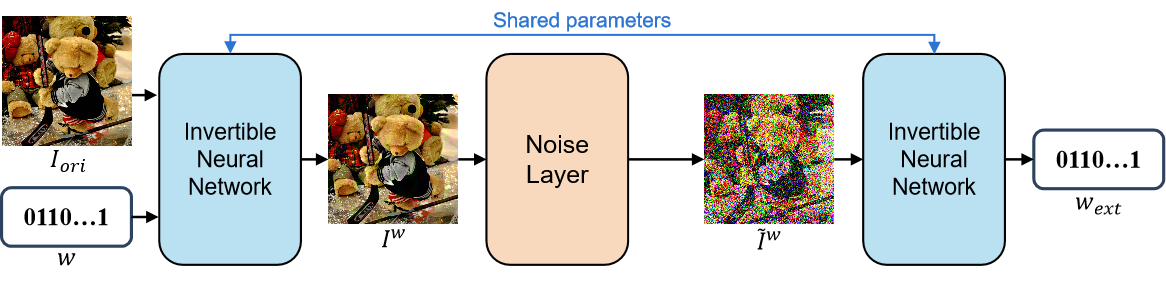}
        \caption{The architecture of flow-based framework.}
        \label{Fig:flowbased}
    \end{subfigure}
    \hfill
    \begin{subfigure}{0.48\textwidth}
        \centering
        \includegraphics[width=\textwidth]{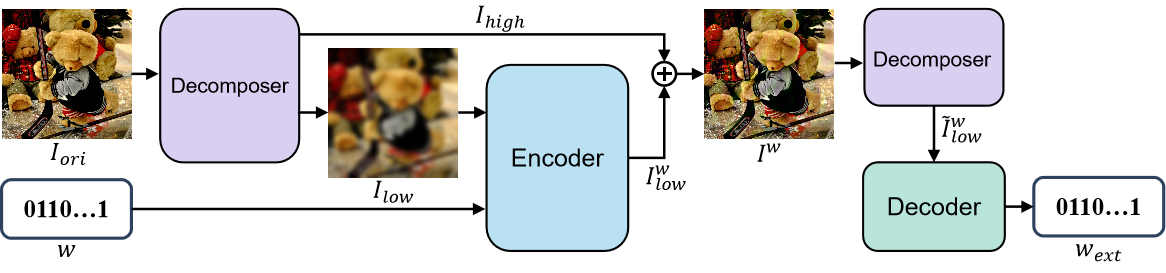}
        \caption{The architecture of our framework.}
        \label{Fig:proposed}
    \end{subfigure}
    \caption{The network comparison of our framework and other frameworks.}
    \label{Fig:framework}
    \vspace{-10pt}
\end{figure}

Current deep learning-based methods primarily improve robustness through noise simulation layers during training. While effective against known attacks, this paradigm heavily relies on pre-defined attack distributions. This approach faces inherent limitations: anticipating all potential distortions is impossible, especially with rapidly evolving AIGC technologies; complex noise layers complicate model design and training; and models overfit the limited simulated attacks, limiting generalization in real-world environments.

To address these limitations, this work explores an alternative path that pivots from attack simulation to exploiting the inherent invariance of images, motivated by the stability of low-frequency components. These components, which encode fundamental structural information, are known to be inherently robust to conventional signal processing~\cite{shijunxiangInvariantImageWatermarking2008}. Crucially, we argue that AIGC models also preserve these low-frequency components to ensure the visual plausibility of their outputs. To build a robust watermark upon this stable foundation, we leverage deep features by integrating the stability prior of low-frequency components with the semantic robustness of deep representations.


Based on this rationale, we propose \textbf{SimuFreeMark}, a noise-\underline{simu}lation-\underline{free} water\underline{mark}ing framework (Fig.~\ref{Fig:proposed}) that symmetrically leverages low-frequency components for both embedding and extraction. The embedding process begins by decomposing the original image via a frequency domain transform to obtain its low-frequency component, which is then encoded by a pre-trained variational autoencoder (VAE). The watermark is embedded into this latent space, where it is bound to the image's structured deep features to enhance semantic robustness. The watermarked latent features are subsequently decoded and fused with the original high-frequency component to reconstruct the final watermarked image. For extraction, the same decomposition yields the low-frequency component, from which the watermark is directly decoded. By eliminating hand-crafted noise layers, our framework derives robustness from the inherent stability of low-frequency components and the representational power of deep features, establishing a simulation-free paradigm for robust watermarking with improved generalization.

The main contributions are summarized as follows.
\begin{itemize}
\item[1)]Through systematic quantitative analysis, we establish that the low-frequency components of images exhibit significant stability against both conventional signal processing and AIGC semantic edits, providing a novel theoretical perspective for watermark design.
\vspace{-4pt}

\item[2)]We propose the SimuFreeMark framework, which eliminates the need for hand-crafted noise simulation by leveraging the inherent image stability to achieve robustness.
\vspace{-4pt}

\item[3)]We introduce a strategy for embedding watermarks in the deep features mapped from low-frequency components, ensuring robustness through the integration of low-frequency stability and deep semantic features.
\vspace{-4pt}

\item[4)]Extensive experiments show that SimuFreeMark outperforms state-of-the-art (SOTA) methods against a wide range of signal processing and AIGC edits, while maintaining superior visual quality.
\end{itemize}

\section{Related Work}
\label{sec:related}
Current deep learning-based watermarking methods can be primarily categorized into the encoder-noise layer-decoder (END) framework and the flow-based framework.

\subsection{END Framework}
The END framework (Fig.~\ref{Fig:endbased}) enables end-to-end training through a pipeline where an encoder embeds watermark, a noise layer simulates attacks to improve robustness, and a decoder extracts watermark from attacked images.

HiDDeN~\cite{zhuHiDDeNHidingData2018} established a standard paradigm for this framework, employing adversarial training to ensure robustness against operations such as JPEG compression and cropping. Subsequent studies developed more sophisticated noise layers for complex scenarios: PIMoG~\cite{fangPIMoGEffectiveScreenshooting2022} simulates perspective distortion and moiré patterns against screen-shooting attacks, while MBRS~\cite{jiaMBRSEnhancingRobustness2021} combines real and simulated JPEG compression in mixed-batch training to optimize robustness against compression distortions.

To counteract AIGC semantic edits, recent END-based methods employ diverse strategies. EditGuard~\cite{zhangEditGuardVersatileImage2024} embeds watermarks in the multi-scale features of the image, but its noise layer design remains limited to conventional noise simulation, offering inadequate robustness against AIGC edits. Robust-Wide~\cite{huRobustWideRobustWatermarking2025} integrates instruction-driven editing models (e.g., InstructPix2Pix~\cite{brooksInstructPix2PixLearningFollow2023}) directly into training, specializing against AIGC edits at the cost of signal processing robustness. VINE~\cite{luRobustWatermarkingUsing2024} uses blurring distortions as noise, motivated by similar frequency-domain effects between AIGC edits and blurring.

The performance of the END framework heavily depends on pre-defined attack distributions, limiting generalization to unseen attacks. Moreover, complex noise layers increase the cost of model design and training.

\subsection{Flow-Based Framework}
The flow-based framework (Fig.~\ref{Fig:flowbased}) utilizes invertible neural networks to construct a bijective mapping between original and watermarked images, ensuring inherent encoding-decoding consistency.

FIN~\cite{fangFlowbasedRobustWatermarking2023} employs an invertible noise layer to simulate conventional signal processing attacks, enhancing robustness while preserving visual quality. CIN~\cite{maBlindWatermarkingCombining2022} combines reversible and irreversible mechanisms: invertible networks for invisibility and an attention mechanism for robustness. IRWArt~\cite{luoIRWArtLeveringWatermarking2023} embeds watermarks in the frequency-domain coefficients of the image for better invisibility and applies adversarial training against frequency distortions.

The flow-based framework ensures consistency between the embedding and extraction processes. However, the complex invertible structure is computationally costly and inherently limits the use of complex noise simulation, restricting its adaptability to various attacks.

\subsection{Summary}
Current deep learning-based watermarking derives robustness from hand-crafted noise layers during training. Whether conventional noise layers in the END framework or invertible noise layers in the flow-based framework, their essence is to guide the model learning by simulating pre-defined attack distributions. Their generalization capability is often insufficient when faced with distortions that are not included in the training process. This limitation motivates the exploration of a new pathway. Instead of attempting to simulate an ever-expanding set of external attacks, we derive robustness directly from images' inherent stable properties, establishing our simulation-free paradigm.

\begin{figure}[!t]
    \centering
    \setlength{\belowcaptionskip}{0pt}
    \includegraphics[width=0.45\textwidth]{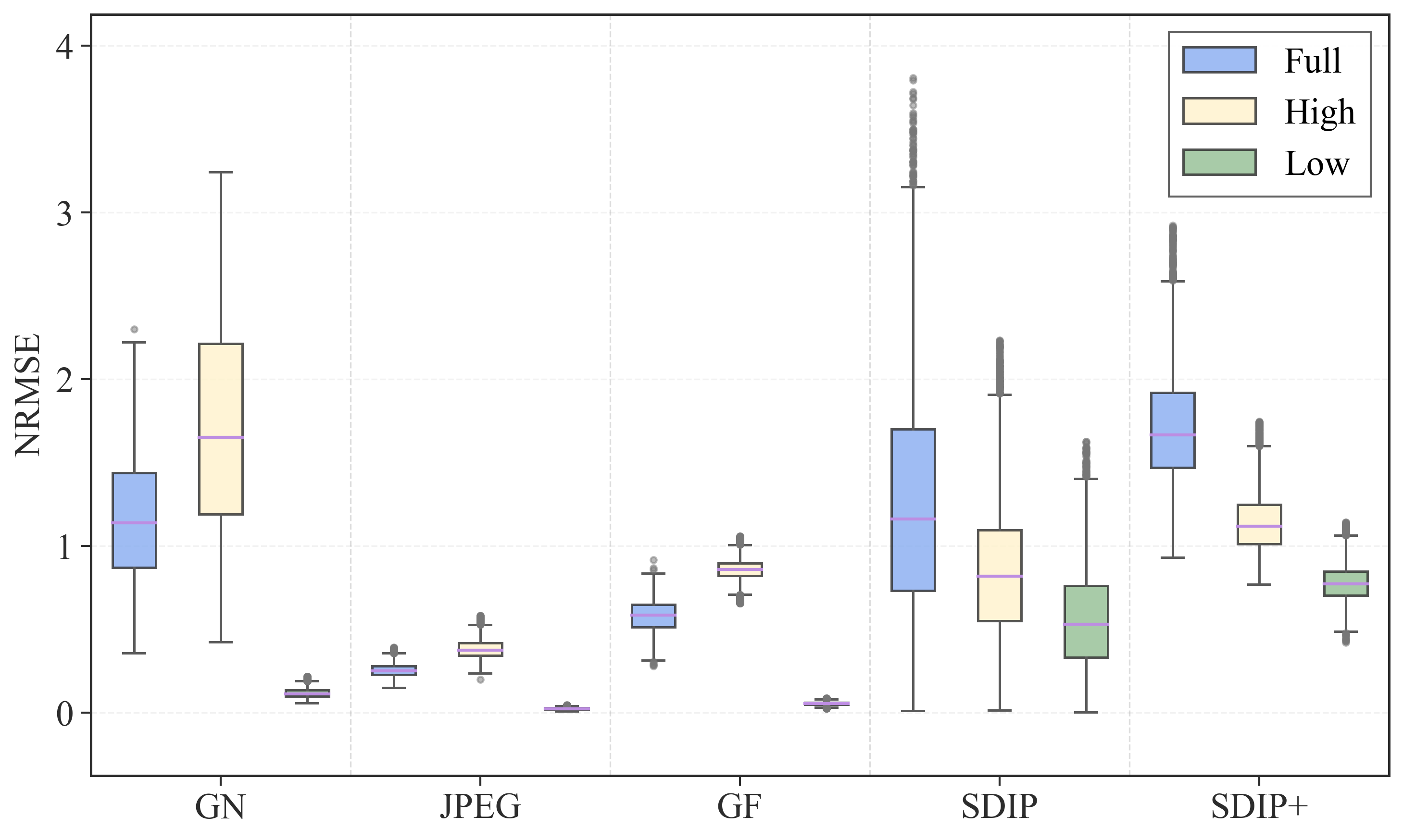}
    \caption{Stability comparison of frequency components.}
    \label{Fig:stability}
\vspace{-10pt}
\end{figure}

\begin{figure*}[t!]
    \centering
    \includegraphics[width=0.98\linewidth]{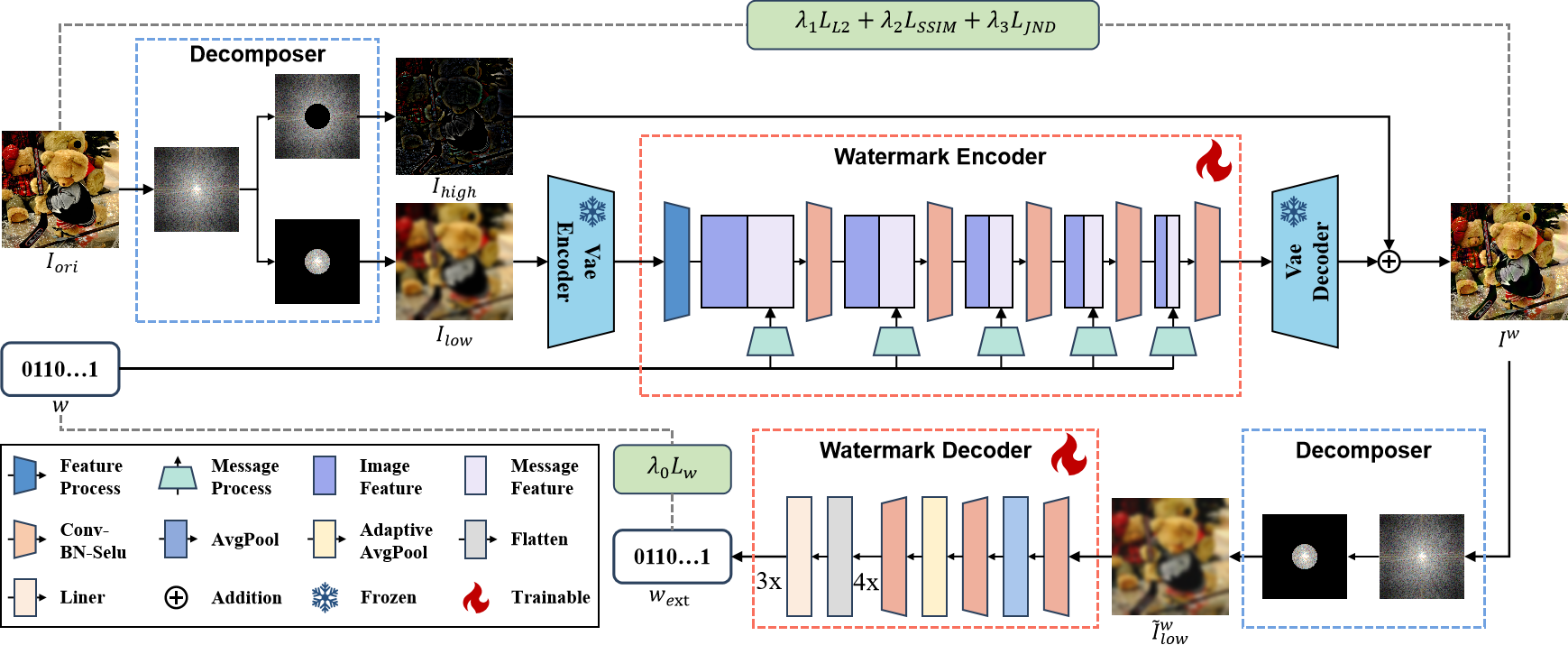}
    \caption{The flowchart of SimuFreeMark.}
    \label{Fig:SimuFreeMark}
\vspace{-10pt}
\end{figure*}

\section{Method}
\label{sec:method}

\subsection{Low-Frequency Stability: Empirical Evidence and Motivations}
\label{subsec:analysis}

\subsubsection{Quantitative Analysis of Stability}
To assess the stability differences across frequency components, we randomly selected 5,000 images from the COCO dataset~\cite{linMicrosoftCOCOCommon2014}. For each image $I$, we applied a set of attacks, including conventional signal processing attacks and AIGC edits, to obtain an attacked image $I'$, with specific parameters: Gaussian noise (GN, standard deviation $\sigma$=0.1), JPEG compression (quality factor $q$=50), Gaussian filtering (GF, kernel size $k$=3$\times$3), Stable Diffusion inpainting (SDIP)~\cite{rombachHighresolutionImageSynthesis2022} (strength $s$=0.7), and Stable Diffusion image-to-image regeneration via inpainting (SDIP+)~\cite{rombachHighresolutionImageSynthesis2022} (strength $s$=0.5).

The stability was quantified in the frequency domain through the following procedure: 
\vspace{-2pt}
\begin{itemize}
\item[1)] Perform 2D Fast Fourier Transform (FFT) on both $I$ and $I'$ to obtain their frequency spectra $F$ and $F'$; 
\vspace{-4pt}
\item[2)] Use a circular low-pass filter $M$ with a radius of 20\% of the central frequency to separate the spectra into low-frequency ($F_{low} = F \odot M$, $F'_{low} = F' \odot M$) and high-frequency components ($F_{high} = F \odot (1-M)$, $F'_{high} = F' \odot (1-M)$), where $\odot$ denotes element-wise multiplication;
\vspace{-4pt}
\item[3)] Compute three normalized root mean square errors (NRMSE) for spectral differences, denoted as $E_{global}$ (global), $E_{low}$ (low-frequency), and $E_{high}$ (high-frequency). As an example, $E_{global}$ is calculated as:
\begin{equation}
      E_{global} = \frac{ \sqrt{ \frac{1}{N} \sum_{i=1}^{N} (F_{i} - F'_{i})^2 } }{ \sqrt{ \frac{1}{N} \sum_{i=1}^{N} (F_{i})^2 } } ,
\end{equation}
where $ N $ is the number of elements in the spectrum. The errors $ E_{low} $ and $ E_{high} $ are computed analogously from their corresponding spectral components, $ F_{low}, F'_{low} $ and $ F_{high}, F'_{high} $.
\end{itemize}
\vspace{-2pt}
Fig.~\ref{Fig:stability} presents the results. For all five attacks, $E_{low}$ is lower than $E_{high}$. This difference is statistically confirmed by paired t-tests on magnitude spectra (p-values $<10^{-4}$ for each attack), demonstrating that low-frequency components exhibit significantly greater stability against these diverse attacks compared with high-frequency components.

\subsubsection{From Empirical Evidence to Framework Design}
The stability of low-frequency components arises because they capture the fundamental structure and global context of an image. Conventional signal processing operations rarely disrupt this structural foundation. Moreover, AIGC models typically retain these components to ensure visual plausibility and structural integrity in their edited outputs.

Building upon this foundation, our framework embeds the watermark into the stable low-frequency region, which derives its robustness from the inherent properties of the image. To further resist semantic-level manipulations, the watermark is embedded not in the pixel-level low-frequency components but in the deep feature space derived from them. This integrates the physical stability of the frequency domain with the semantic robustness of deep representations, forming the core of our framework.

\subsection{The SimuFreeMark Framework}
This section details the overall architecture of the SimuFreeMark framework, as illustrated in Fig.~\ref{Fig:SimuFreeMark}.

\subsubsection{Watermark Embedding}
The embedding process consists of three steps: original image decomposition, latent space embedding, and watermarked image reconstruction.

\textbf{Original image decomposition:} The original image $I_{ori} \in \mathbb{R}^{3 \times H \times W}$ is transformed by FFT to obtain its frequency spectrum, which is separated by a circular low-pass filter that preserves the central 20\% of frequency components into low-frequency ($F_{low}$) and high-frequency ($F_{high}$) components. Inverse FFT reconstructs these components into spatial-domain images $I_{low}$ and $I_{high}$.

\textbf{Latent space embedding:} The low-frequency image $I_{low}$ is encoded by a frozen, pre-trained Stable Diffusion VAE (SD-VAE) encoder~\cite{rombachHighresolutionImageSynthesis2022} into a latent representation $Z_{latent} \in \mathbb{R}^{4 \times H' \times W'}$, where $(H',W')$ denote latent space dimensions. Concurrently, the watermark sequence $w \in \{0,1\}^L$ (of length $L$) is normalized to $[-0.5, 0.5]$. Both inputs feed into a two-branch embedding module:

\vspace{-4pt}
\begin{itemize}
\item The watermark branch first transforms the normalized sequence via two fully-connected layers (with SELU activations~\cite{klambauerSelfNormalizingNeuralNetworks2017}) to expand its dimension to 4096. This vector is then reshaped, processed by a 1×1 convolution, and fed through four Conv-BN-SELU blocks (each a 3×3 convolution with stride 1 and padding 1, followed by Batch Normalization and SELU) with output channels of 128, 32, 16, and 8, respectively.
\vspace{-4pt}

\item The latent feature branch processes \( Z_{latent} \) through a single Conv-BN-SELU block, projecting its 4 input channels to 128 output channels.
\end{itemize}
\vspace{-4pt}

The feature fusion employs a multi-stage design. The 128-channel outputs from both branches are first concatenated. This combined tensor and the evolving watermark features then undergo four parallel fusion stages. In each stage, both streams are processed by separate Conv-BN-SELU blocks that halve their channel dimensions, and their outputs are concatenated to form the input for the subsequent stage. This process progressively reduces channel counts while deepening feature integration.

The final outputs from both paths, each with 4 channels, are concatenated and passed through a final 1×1 convolution to produce a residual feature map \( Z_{fuse} \). This residual is scaled by a factor of \( \alpha = 0.2 \) and added to the original latent to produce the watermarked latent feature:
\begin{equation}
    Z_{latent}^w = Z_{latent} + \alpha Z_{fuse}.
\end{equation}

$Z_{latent}^w$ is then decoded by the frozen SD-VAE decoder to produce the watermarked low-frequency image $I_{low}^w$.

\textbf{Watermarked image reconstruction:} The final watermarked image $I^w$ is reconstructed by combining $I_{low}^w$ with the original high-frequency component $I_{high}$, followed by clamping to ensure it lies within the valid pixel range:
\begin{equation}
    I^w = \text{clamp}(I_{low}^w + I_{high}, 0, 1).
\end{equation}

\subsubsection{Watermark Extraction}
The extraction process directly recovers the watermark from the low-frequency component of the watermarked image $I^w$, and consists of two steps.

\textbf{Watermarked image decomposition:} The watermarked image $I^w$ undergoes the same frequency domain decomposition and reconstruction process as the embedding stage to obtain its low-frequency component image $\tilde{I}^{w}_{low}$.

\textbf{Watermark decoding:} The image $\tilde{I}^{w}_{low}$ is then fed into a watermark extraction network. 

The network begins with a feature extraction module comprising two Conv-BN-SELU blocks. The first block increases the input channels from 3 to 256, and the second reduces the channels to 128. A 2×2 average pooling layer is applied after the first block, halving the spatial dimensions.

The resulting feature map is then adaptively averaged pooled to a fixed size of \(64 \times 64\) to ensure dimensional consistency for subsequent layers. This is followed by a decoding module composed of four Conv-BN-SELU blocks that progressively reduce the channel dimensions from 128 to 32, 16, 8, and finally 4. A final 1×1 convolution reduces the channel dimension to 1, after which the feature map is flattened into a vector.

This vector is processed by a series of three fully-connected layers (with SELU activations for the first two) that gradually reduce the dimension to the target watermark length \(L\). The final \(L\)-dimensional output vector is passed through a sigmoid function to produce the extracted watermark probabilities, which are then binarized via rounding to obtain the final watermark sequence:
\begin{equation}
w_{ext} = \text{round}(\text{sigmoid}(f_{decoder}(\tilde{I}^w_{low}))),
\end{equation}
where $f_{decoder}$ represents the entire extraction network.

\subsection{Objective Function and Training Strategy}
To ensure accurate watermark extraction while maintaining high visual quality, we designed a compound objective function and adopted a two-stage training strategy to balance the model's optimization direction.

\subsubsection{Objective Function}
The training of SimuFreeMark is guided by a compound objective function designed to balance watermark robustness with visual quality. The total loss $\mathcal{L}_{total}$ is a weighted sum of four components:
\begin{equation}
\mathcal{L}_{total} = \lambda_0 \mathcal{L}_w + \lambda_1 \mathcal{L}_{L2} + \lambda_2 \mathcal{L}_{SSIM} + \lambda_3 \mathcal{L}_{JND}.
\end{equation}

Here, \(\mathcal{L}_w = \text{BCE}(w, w_{\text{ext}})\) is the binary cross-entropy loss that serves as the primary driver for extraction accuracy. For visual quality, the L2 loss \(\mathcal{L}_{L2} = \| I_{\text{ori}} - I^w \|_2^2\) enforces pixel-level consistency, while the structural similarity loss \(\mathcal{L}_{\text{SSIM}} = 1 - \text{SSIM}(I_{\text{ori}}, I^w)\) preserves the image's global structure. Finally, perceptual imperceptibility is achieved through a just noticeable difference (JND) loss~\cite{ganGenPTWIngenerationImage2025}, \(\mathcal{L}_{\text{JND}} = (1 - 2 \cdot \text{JND}(I_{\text{ori}})) \cdot | I_{\text{ori}} - I^{w} |\), which adaptively confines embedding distortions below the visual sensitivity threshold by leveraging a spatial JND map. Specifically, the JND map $\text{JND}(I_{\text{ori}})$ is computed following the model proposed by~\cite{wuEnhancedJustNoticeable2017}.

\subsubsection{Training Strategy}
\label{subsubsec:train_stra}
We employ a two-stage training strategy to balance robustness and visual quality. The first stage prioritizes building strong robustness against various attacks. The loss function is configured to favor extraction accuracy, with the following weights: $\lambda_0 = 5$, $\lambda_1 = 0.1$, $\lambda_2 = 5$, $\lambda_3 = 10$. Training proceeds until the watermark extraction loss $\mathcal{L}_w$ converges below a threshold of 0.05, indicating that the model has acquired reliable extraction capability.

The second stage aims to enhance visual quality while preserving the robustness achieved in the first stage. To impose stronger constraints on image fidelity, we adjust the loss weights as follows: $\lambda_0 = 5$, $\lambda_1 = 1$, $\lambda_2 = 10$, $\lambda_3 = 50$. By increasing the weights of the penalties for pixel-level, structural, and perceptual distortions, this stage guides the model to generate watermarked images that are visually closer to the originals.

With the training configuration detailed in Section~\ref{subsec:setup}, the first stage typically converged within 3,000 iterations. The second stage was then activated for visual refinement.

\begin{table*}[!t]
\centering
\small
\caption{Robustness of compared methods in bit accuracy (\%) under conventional signal processing attacks.}
\label{tab:comparison_csp}
    \vspace{-3pt}
\begin{tabular}{lccccccccccc}
\toprule
\multirow{2}{*}{Method}	&\multirow{2}{*}{PSNR/SSIM} &\multicolumn{3}{c}{GN} &\multicolumn{3}{c}{S\&PN}  &\multicolumn{4}{c}{JPEG}\\
\cmidrule(lr){3-5}	\cmidrule(lr){6-8}	\cmidrule(lr){9-12}
&   &$\sigma$=0.1	&0.15	&0.2	&$d$=0.1	&0.15	&0.2	&$q$=10	&30	&50 &70\\
\midrule
HiDDeN & 27.62/0.897 &  51.18&	50.76&	50.37&    59.86&	55.20&	53.09&
50.16&	50.10&	50.48&	50.89\\ 
PIMoG & 35.94/0.891 & 64.61&	59.08&	56.29&  \underline{98.88}&	\underline{97.13}&	\underline{94.83}&    53.58&	68.51&	79.61&	89.69\\
RoSteALS & 32.35/0.953 & \underline{96.61}&	\underline{94.43}&	\underline{92.16}&	96.85&	95.67&	94.35&  90.17&	97.04&	98.22&	98.81 \\
FIN & 38.18/0.981 	&92.06&	82.75&	75.83&	72.59&	68.05&	65.13&	82.55&	\textbf{99.86}&	\textbf{100.00}&	\textbf{100.00}\\
EditGuard & 37.12/0.902 &74.02&	66.64&	62.72&	98.16&	96.41&	93.22&	62.98&	73.31&	83.42&	88.92\\
Robust-Wide & 39.18/0.905 & 95.05&	88.30&	81.51&	98.45&	92.09&	82.17&	\textbf{98.23}&	\textbf{99.86}&	\textbf{100.00}&	\textbf{100.00}\\
SimuFreeMark & \textbf{39.19/0.990} & \textbf{99.66}&	\textbf{99.14}&	\textbf{99.20}&	\textbf{99.89}&	\textbf{99.90}&	\textbf{99.86}&	\underline{94.31}&	\underline{99.22}&	\underline{99.81}&	\underline{99.77}\\
\midrule
\multirow{2}{*}{Method}	&\multirow{2}{*}{PSNR/SSIM} &\multicolumn{2}{c}{Contr}  &\multicolumn{2}{c}{Bright} &\multicolumn{2}{c}{GF}   &\multicolumn{2}{c}{MeanF}   &\multicolumn{2}{c}{MedF}\\
\cmidrule(lr){3-4}	\cmidrule(lr){5-6}	\cmidrule(lr){7-8}	\cmidrule(lr){9-10} \cmidrule(lr){11-12}
&   &$r$=$\pm$20	&$\pm$40	&$r$=$\pm$15	&$\pm$30	&$k$=5	&7	&$k$=5	&7	&$k$=5	&7	\\
\midrule
HiDDeN & 27.62/0.897 &98.90	&98.26&	98.95&	98.50&	53.29&	50.94&	52.75&	51.49&	51.62&	51.20  \\ 
PIMoG & 35.94/0.891  &\textbf{99.90}	&99.62&	99.82&	99.47&	99.88&	99.74&	99.73&	98.76&	99.81&	98.91  \\
RoSteALS & 32.35/0.953 & 98.88&	98.34&  98.92&	98.50&   99.23&	99.20&   99.20&	99.16&	99.22&	99.17  \\
FIN & 38.18/0.981 	&99.50&	98.06&	98.91&	98.38&	88.02&	80.75&	45.14&	43.52&	59.95&	58.14\\
EditGuard & 37.12/0.902 &\underline{99.89}	&\underline{99.84}&	\textbf{99.94}&	99.52&	87.88&	83.27&	68.53&	85.08&	62.88&	70.83\\
Robust-Wide & 39.18/0.905 &\textbf{99.90}	&99.78&	\underline{99.86}&	\underline{99.61}&	\textbf{99.98}&\underline{99.85}&	\textbf{99.99}&	\underline{99.87}&	\underline{99.89}&	\underline{99.82}\\
SimuFreeMark & \textbf{39.19/0.990} &99.88	&\textbf{99.94}&	\textbf{99.94}&	\textbf{99.84}&	\underline{99.91}&	\textbf{99.86}&	\underline{99.92}&	\textbf{99.89}&	\textbf{99.94}&	\textbf{99.92}\\
\bottomrule
\end{tabular}
\end{table*}

\section{Experiments}
\label{sec:exper}
\subsection{Experimental Setup}
\label{subsec:setup}
\textbf{Training configuration:} SimuFreeMark is trained on a randomly sampled subset of 20,000 images from the COCO training dataset~\cite{linMicrosoftCOCOCommon2014}, with all images resized to $512\times512$. The model is trained using the AdamW optimizer with a learning rate of \( 1 \times 10^{-4} \), a batch size of 2, and a weight decay of \( 1 \times 10^{-2} \). All experiments are conducted on an NVIDIA RTX 3090 GPU. The pre-trained SD-VAE model from~\cite{rombachHighresolutionImageSynthesis2022} is used and kept frozen throughout the training.

\textbf{Baselines and evaluation:} We compare SimuFreeMark with four robust watermarking schemes: HiDDeN~\cite{zhuHiDDeNHidingData2018}, PIMoG~\cite{fangPIMoGEffectiveScreenshooting2022}, RoSteALS~\cite{buiRoSteALSRobustSteganography2023}, FIN~\cite{fangFlowbasedRobustWatermarking2023}, EditGuard~\cite{zhangEditGuardVersatileImage2024}, and Robust-Wide~\cite{huRobustWideRobustWatermarking2025}. For fair comparison, all methods are evaluated with a fixed watermark length of 64 bits. The models for HiDDeN, PIMoG, and FIN were adapted to this message length by retraining using their official codebases, while EditGuard and Robust-Wide are used with their released models that natively embed 64-bit messages. The evaluation uses 1,000 randomly selected images from the UltraEdit dataset~\cite{zhaoUltraEditInstructionbasedFinegrained2024}. The AIGC semantic edits adhere to the UltraEdit protocol, using the source images, editing instructions, and object masks provided by it.

\textbf{Metrics:} Visual quality is measured by PSNR and SSIM (higher values indicate better invisibility). Robustness is evaluated using the average bit accuracy, defined as the proportion of correctly extracted watermark bits.

\begin{table*}[!t]
\centering
\small
\caption{Robustness of compared methods in bit accuracy (\%) under AIGC semantic edits.}
\label{tab:comparison_aigc}
    \vspace{-3pt}
\begin{tabular}{lcccccccccc}
\toprule
\multirow{3}{*}{Method}	&\multirow{3}{*}{PSNR/SSIM} &\multicolumn{5}{c}{Local Edit}  &\multicolumn{4}{c}{Global Edit}\\
\cmidrule(lr){3-7}	\cmidrule(lr){8-11}
&   &\multirow{2}{*}{LaMa} &\multirow{2}{*}{RD}  &\multicolumn{3}{c}{SDIP}    &\multirow{2}{*}{P2P}   &\multicolumn{3}{c}{SDIP+}\\
\cmidrule(lr){5-7}  \cmidrule(lr){9-11}
&   &	&&$s$=0.3	&0.5	&0.7	&&$s$=0.1	&0.3	&0.5\\
\midrule
HiDDeN & 27.62/0.897 &  91.78&	92.12&	91.75&	91.67&	91.81&	52.95&	55.78&	54.79&	53.42\\ 
PIMoG & 35.94/0.891 & 92.34&	91.52&	96.04&	93.51&	91.93&	88.36&	93.76&	72.52&	60.68\\
RoSteALS & 32.35/0.953 & 92.53& 91.99&  94.56&	93.38&	92.72&    \underline{92.69}&  83.74&	67.94&	59.84  \\
FIN & 38.18/0.981 	&81.16&	86.41&	81.55&	82.83&	83.52&	51.84&	54.05&	54.42&	54.47\\
EditGuard & 37.12/0.902 &   \underline{92.72}&	\textbf{94.00}&	95.66&	94.55&	\textbf{95.61}&	50.14&	49.31&	50.61&	50.25\\
Robust-Wide & 39.18/0.905 & 92.37&	91.92&	\underline{96.93}&	\underline{94.77}&	93.38&	\textbf{97.36}&	\underline{95.57}&	\underline{78.04}&	\underline{64.30}\\
SimuFreeMark & \textbf{39.19/0.990} &\textbf{94.22}&	\underline{93.86}&	\textbf{97.56}&	\textbf{95.61}&	\underline{94.02}&	90.42&	\textbf{97.64}&	\textbf{94.05}&	\textbf{92.25}\\
\bottomrule
\end{tabular}
\end{table*}

\begin{figure*}[!t]
\centering
\setlength{\tabcolsep}{2pt} 
\renewcommand{\arraystretch}{1} 
\begin{tabular}{cccccccccc}
 \includegraphics[width=0.09\textwidth]{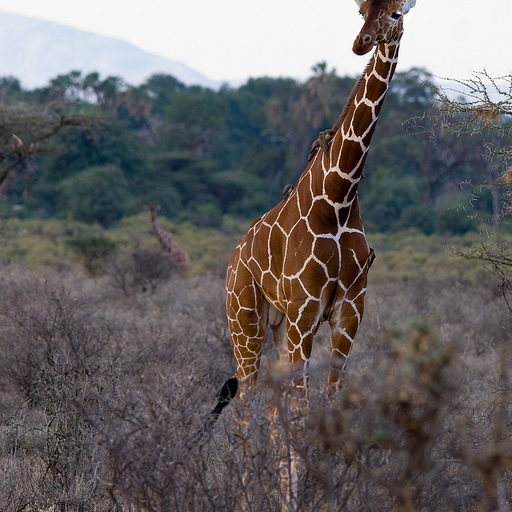} & 
 \includegraphics[width=0.09\textwidth]{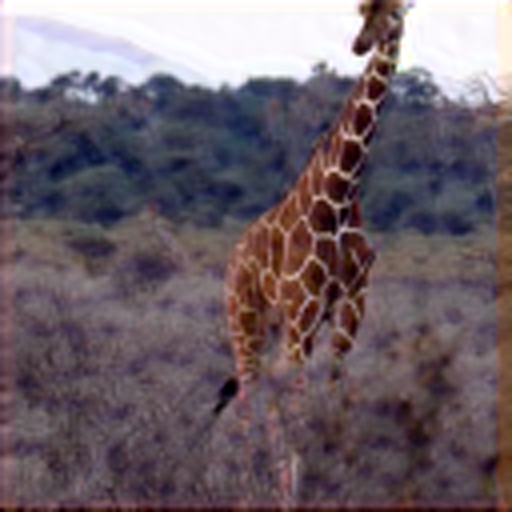} &
 \includegraphics[width=0.09\textwidth]{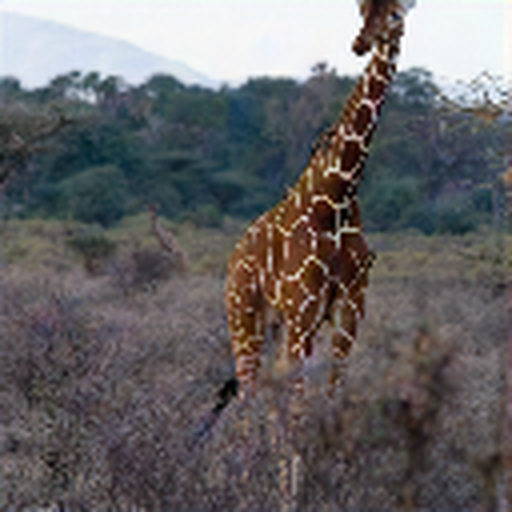} & 
 \includegraphics[width=0.09\textwidth]{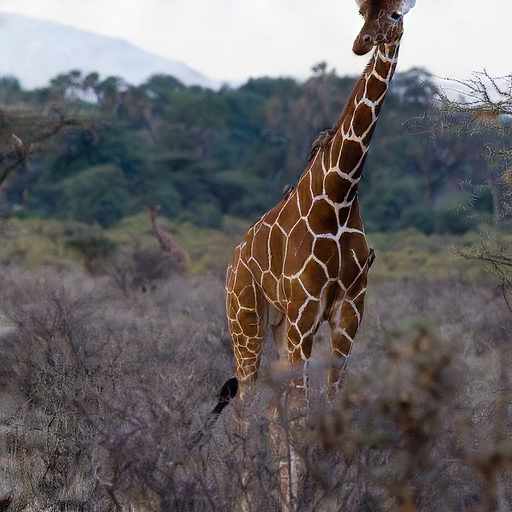} & 
 \includegraphics[width=0.09\textwidth]{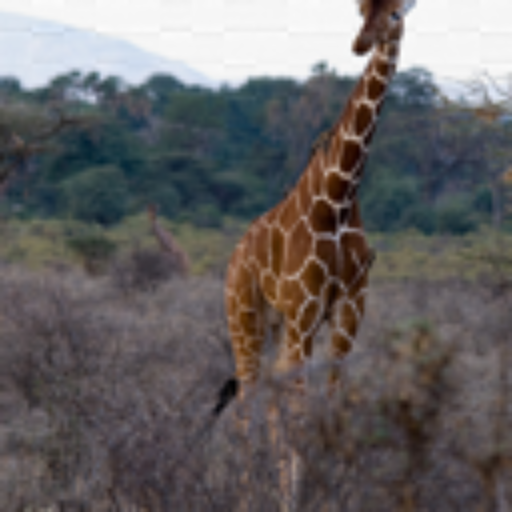} & 
 \includegraphics[width=0.09\textwidth]{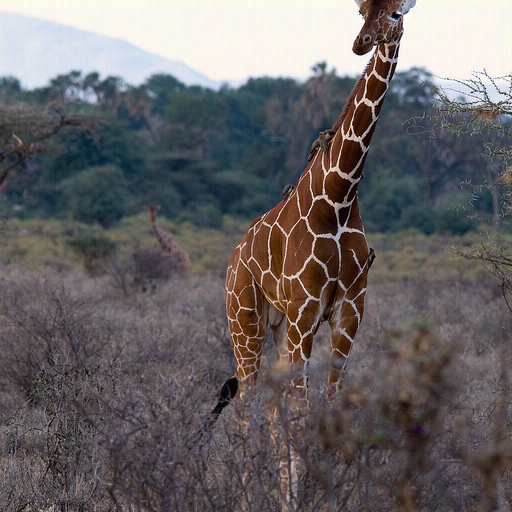} & 
 \includegraphics[width=0.09\textwidth]{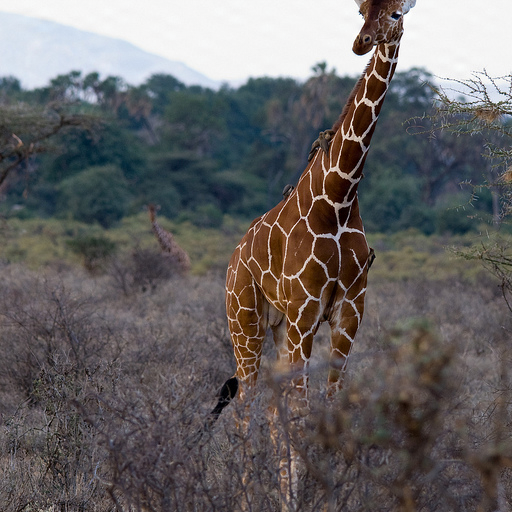} & 
 \includegraphics[width=0.09\textwidth]{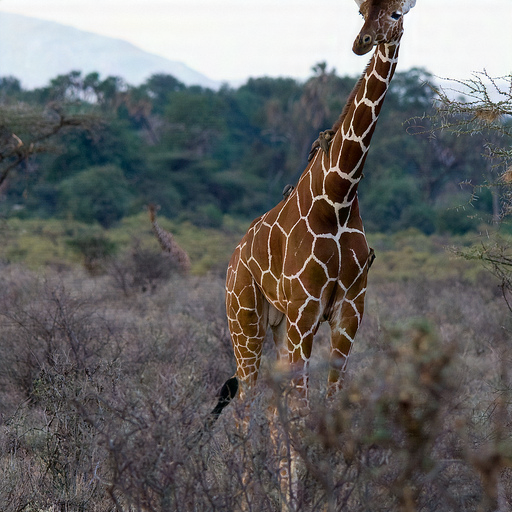}\\
 & 
 \includegraphics[width=0.09\textwidth]{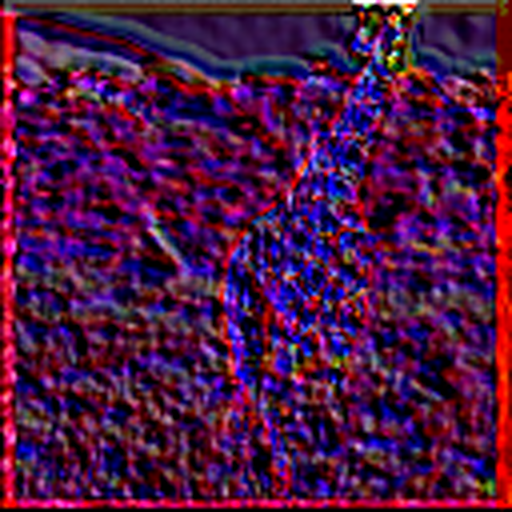} & 
\includegraphics[width=0.09\textwidth]{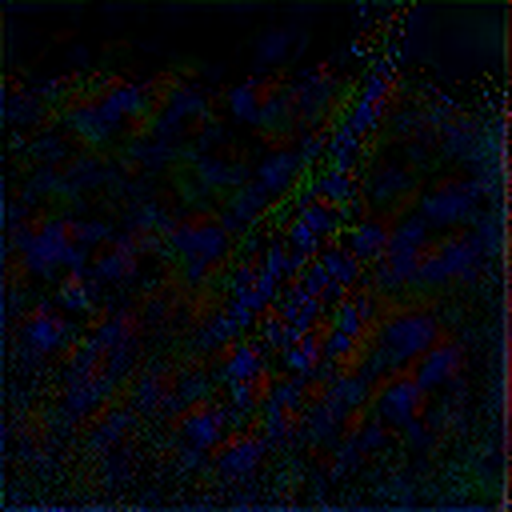} & 
 \includegraphics[width=0.09\textwidth]{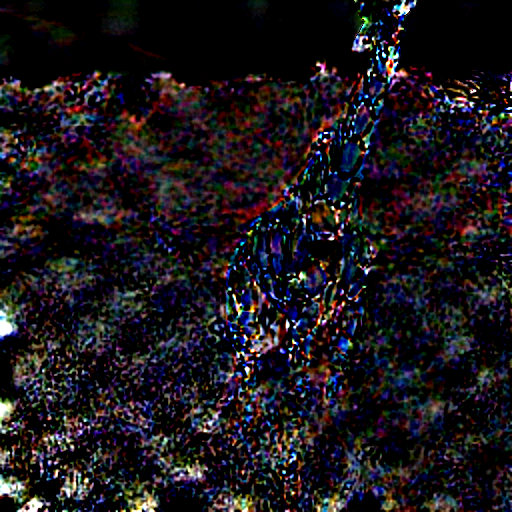} & 
\includegraphics[width=0.09\textwidth]{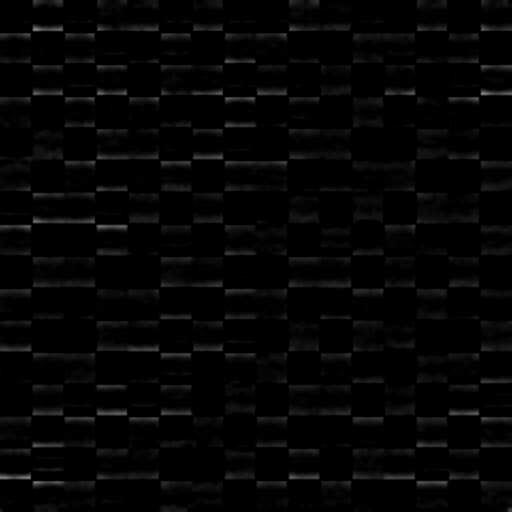} & 
\includegraphics[width=0.09\textwidth]{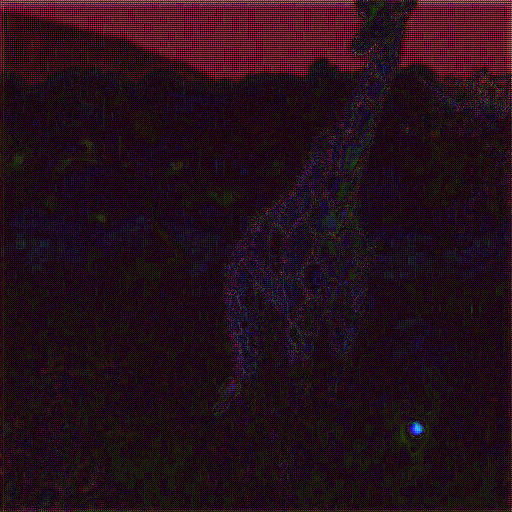} &  
\includegraphics[width=0.09\textwidth]{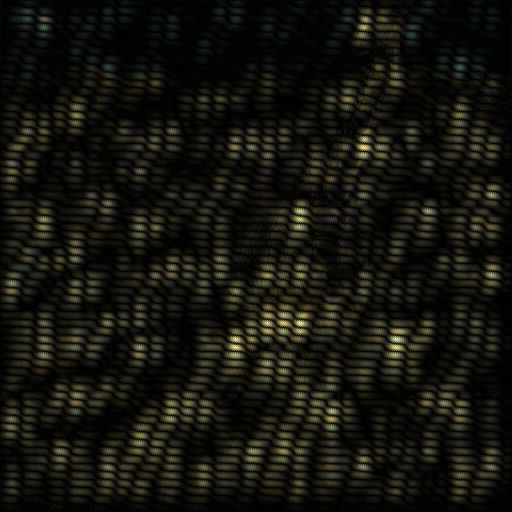} & 
\includegraphics[width=0.09\textwidth]{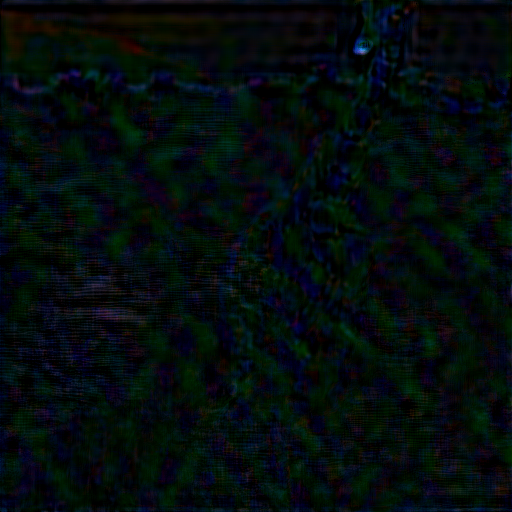}\\

\includegraphics[width=0.09\textwidth]{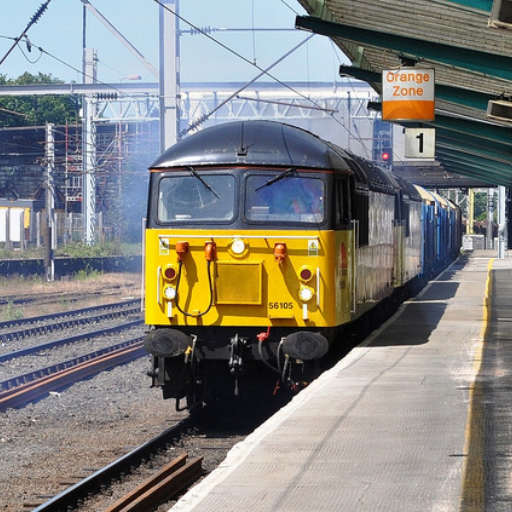} & 
\includegraphics[width=0.09\textwidth]{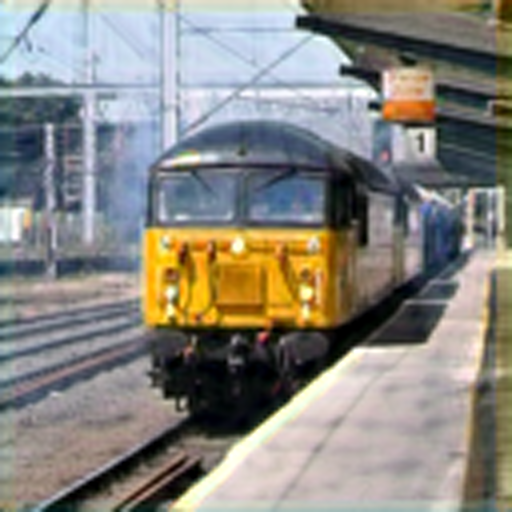} & 
\includegraphics[width=0.09\textwidth]{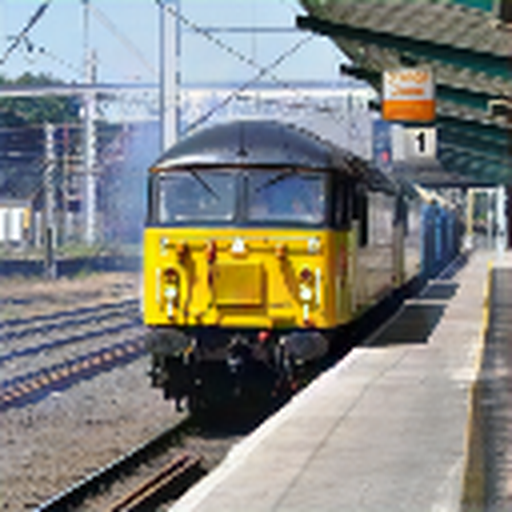} & 
 \includegraphics[width=0.09\textwidth]{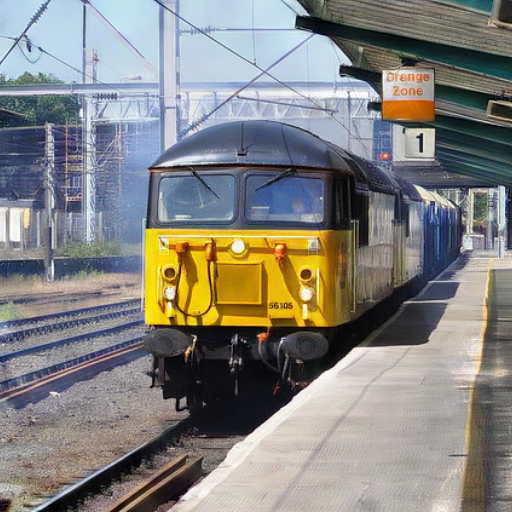} & 
\includegraphics[width=0.09\textwidth]{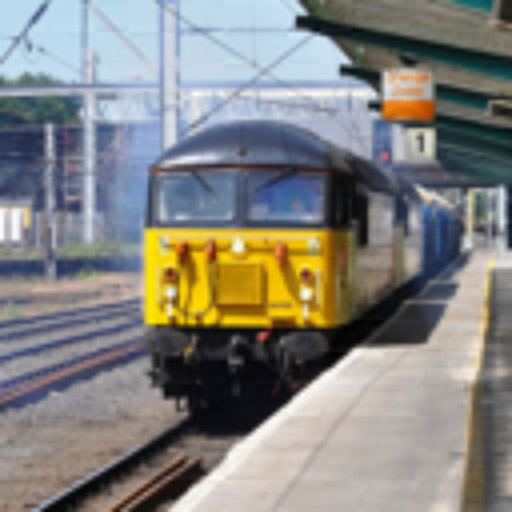} & 
\includegraphics[width=0.09\textwidth]{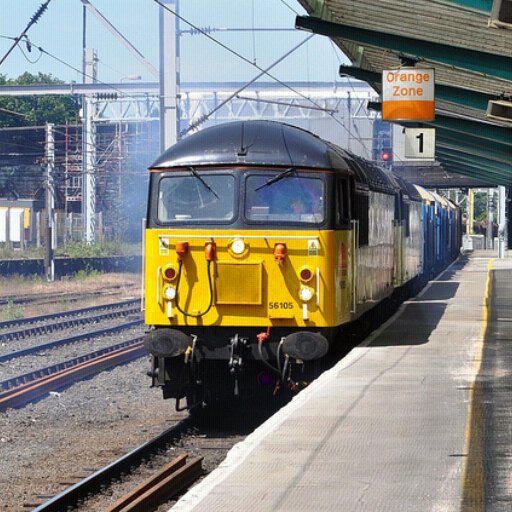} & 
\includegraphics[width=0.09\textwidth]{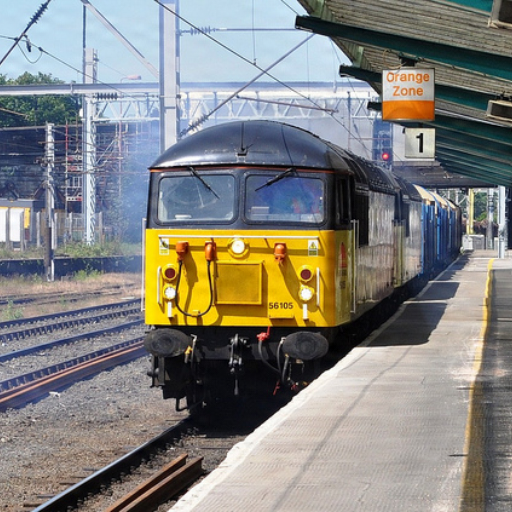} & 
\includegraphics[width=0.09\textwidth]{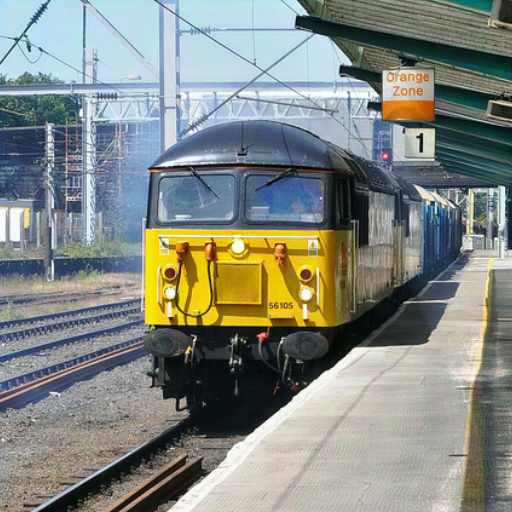}\\
 & 
 \includegraphics[width=0.09\textwidth]{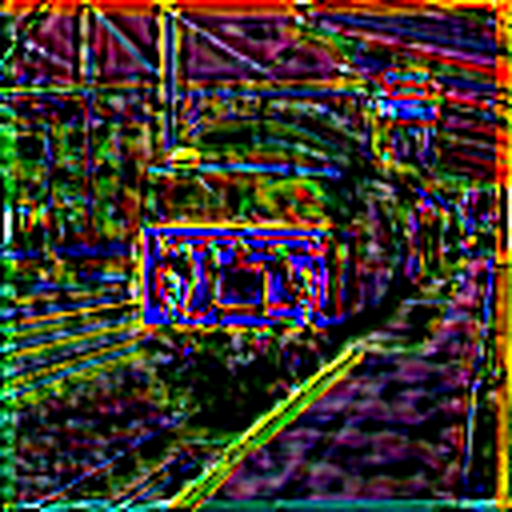} & 
 \includegraphics[width=0.09\textwidth]{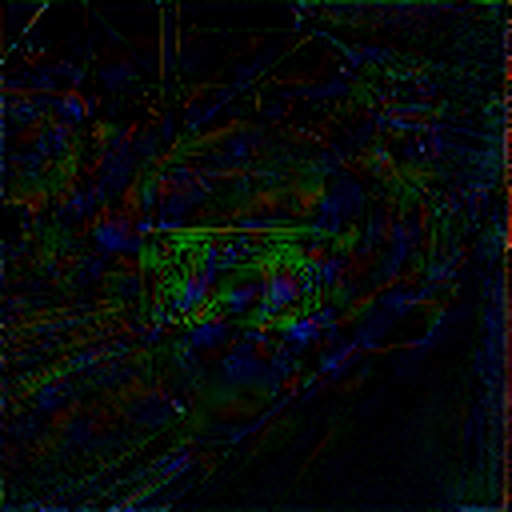} &
 \includegraphics[width=0.09\textwidth]{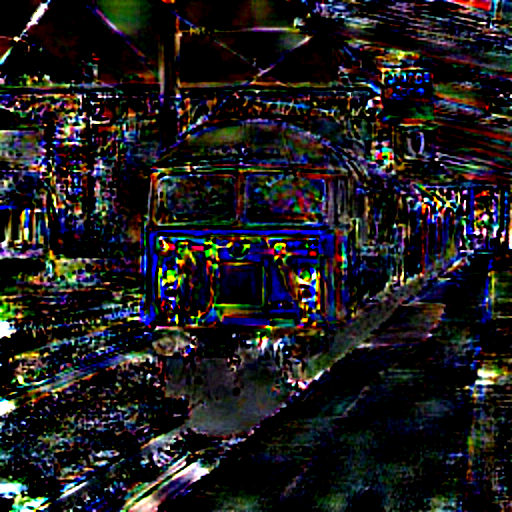} & 
 \includegraphics[width=0.09\textwidth]{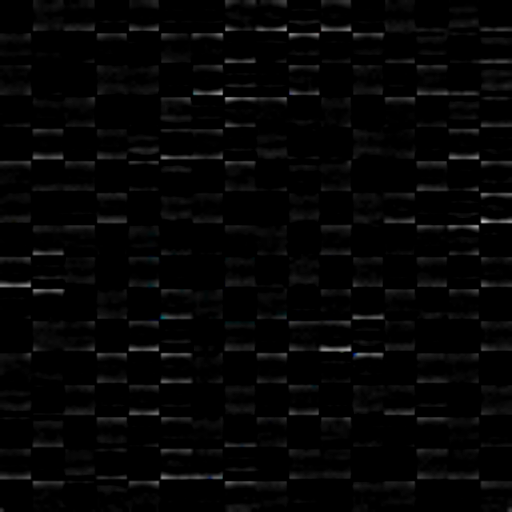} & 
 \includegraphics[width=0.09\textwidth]{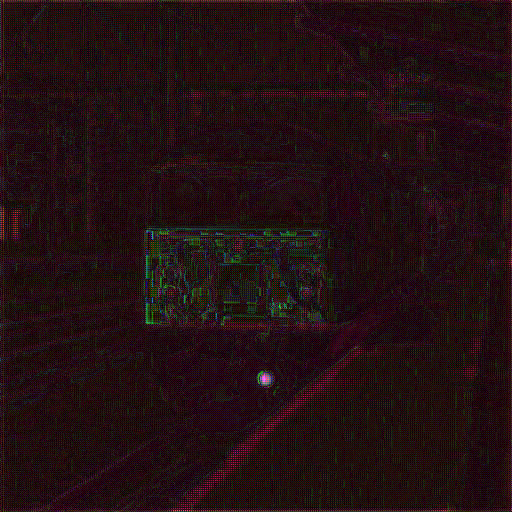} & 
 \includegraphics[width=0.09\textwidth]{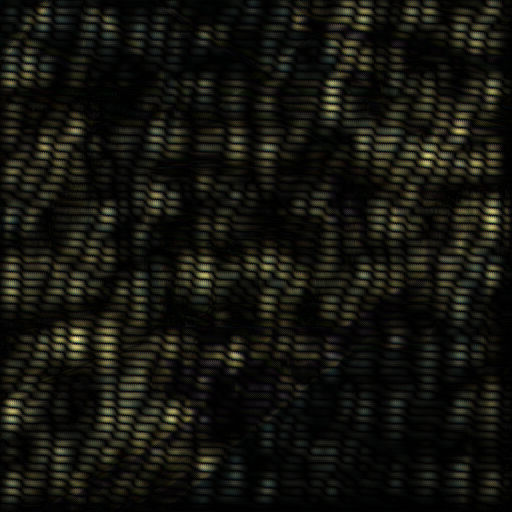} & 
 \includegraphics[width=0.09\textwidth]{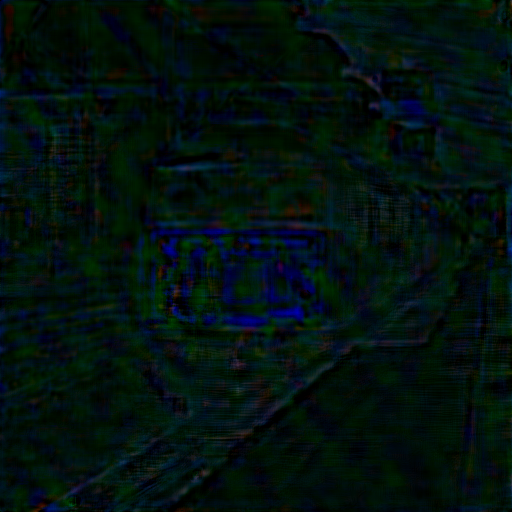}\\
\scriptsize Original Image & \scriptsize HiDDeN & \scriptsize PIMoG  & \scriptsize RoSteALS & \scriptsize FIN & \scriptsize EditGuard & \scriptsize Robust-Wide&\scriptsize Ours
\end{tabular}
\caption{Visual quality comparison with SOTA methods.}
\label{fig:visual}
\vspace{-10pt}
\end{figure*}

\subsection{Comparison with SOTA Methods}
\label{subsec:compare}
This section presents a comprehensive comparison between the SimuFreeMark and SOTA methods. The evaluation covers robustness against both conventional signal processing attacks and AIGC semantic edits, as well as an assessment of the visual quality of the watermarked images.

\subsubsection{Robustness Against Conventional Signal Processing Attacks}
We assess the robustness against eight types of conventional signal processing attacks in Table~\ref{tab:comparison_csp}.

\textbf{Noise:} Under Gaussian noise (GN) (standard deviations $\sigma$ of 0.1, 0.15, and 0.2) and salt-and-pepper noise (S\&PN) (densities $d$ of 0.1, 0.15, and 0.2), SimuFreeMark maintains high accuracy ($\geq$99.14\%), outperforming methods like HiDDeN and PIMoG. This is attributed to the inherent stability of low-frequency components, in which the watermark is less susceptible to pixel-level perturbations.

\textbf{Compression:} SimuFreeMark maintains high robustness across JPEG compression quality factors $q$ ranging from 10 to 70, achieving 94.31\% at $q$=10 and over 99.22\% for $q$$\geq$30. This strong performance occurs because JPEG compression preserves low-frequency information to maintain basic image structure, thereby retaining our embedded watermark. While FIN and Robust-Wide shows competitive performance in compression, SimuFreeMark provides more balanced robustness across various attacks.

\textbf{Photometric adjustments:} Against contrast adjustments (Contr) (factor $r$=$\pm$20\%, $\pm$40\%) and brightness adjustments (Bright) (factor $r$=$\pm$15\%, $\pm$30\%), all methods achieve high accuracy, indicating that current watermarking techniques are robust against these global transformations.

\textbf{Filtering:} For Gaussian filtering (GF), mean filtering (MeanF), and median filtering (MedF) with kernel sizes $k$ of 5$\times$5 and 7$\times$7, SimuFreeMark achieves accuracy above 99.86\%. Since these filtering operations suppress high frequencies, the low-frequency components that host the watermark remain largely unchanged, ensuring its survival.

\textbf{Summary:} The results on conventional attacks validate the premise that the low-frequency component of an image provides a naturally robust carrier. By leveraging this inherent stability, SimuFreeMark achieves superior performance without relying on noise simulation during training.

\subsubsection{Robustness Against AIGC Semantic Edits}
We further evaluate the robustness against AIGC semantic editing attacks, which are categorized into local and global edits. The results are presented in Table~\ref{tab:comparison_aigc}.

\textbf{Local edits:} For local edits including LaMa inpainting~\cite{suvorovResolutionrobustLargeMask2022}, random dropout (RD) of 10-30$\%$ image regions, and SD inpainting (SDIP)~\cite{rombachHighresolutionImageSynthesis2022} with strengths $s$ of 0.3, 0.5, and 0.7, SimuFreeMark achieves competitive performance. It attains the highest accuracy on LaMa (94.22\%) and on SDIP at $s=0.3$ (97.56\%) and $s=0.5$ (95.61\%), demonstrating strong resistance to local content modifications.

\textbf{Global edits:} For global edits, SimuFreeMark shows particularly strong performance against SD image-to-image regeneration via inpainting (SDIP+) across all strength levels $s$=0.1, 0.3, 0.5, achieving the highest accuracy. This superior performance can be explained by the tendency of AIGC models to preserve the low-frequency structure and global composition of the source image during regeneration processes. For InstructPix2Pix (P2P)~\cite{brooksInstructPix2PixLearningFollow2023} instruction-based editing, SimuFreeMark achieves 90.42\% accuracy, showing room for improvement in handling semantic rewriting guided by explicit instructions.

\textbf{Summary:} The experiments on AIGC edits demonstrate that SimuFreeMark provides effective protection against diverse semantic manipulations, with particularly strong performance in regeneration-based attacks. This validates that the use of the stability of low-frequency components offers a promising approach to resist AIGC edits without specialized simulation.

\begin{table*}[!t]
\centering
\small
\caption{Ablation study on low-frequency extractors.}
\label{tab:comparison_extra_method}
    \vspace{-3pt}
\begin{tabular}{lccccccccccc}
\toprule
\multirow{2}{*}{Method}	&\multirow{2}{*}{PSNR/SSIM} &\multicolumn{5}{c}{Conventional Attack}    &\multicolumn{5}{c}{AIGC Semantic Edit}\\
\cmidrule(lr){3-7}	\cmidrule(lr){8-12}
&&GN &JPEG   &Contr   &Bright &GF   &LaMa   &RD   &SDIP &P2P    &SDIP+\\
\midrule
DCT & 38.95/0.988 	&98.56&	\textbf{96.45}&	99.57&	99.50&	99.67&	90.52&	88.31&	92.08&	94.06&	\textbf{94.11}\\
GAU & 39.06/\textbf{0.991} 	&98.53&	93.90&	99.56&	99.78&	\textbf{99.87}&	92.72&	90.14&	92.88&	\textbf{94.25}&	91.20\\
\textbf{FFT} & \textbf{39.19}/0.990 	&\textbf{99.20}&	94.31&	\textbf{99.94}&	\textbf{99.84}&	99.86&	\textbf{94.22}&	\textbf{93.86}&	\textbf{94.02}&	90.42&	92.25\\
\bottomrule
\end{tabular}
\end{table*}

\begin{table*}[!t]
\centering
\small
\caption{Ablation study on the pre-trained VAE.}
\label{tab:comparison_vae}
    \vspace{-3pt}
\begin{tabular}{lccccccccccc}
\toprule
\multirow{2}{*}{Method}	&\multirow{2}{*}{PSNR/SSIM} &\multicolumn{5}{c}{Conventional Attack}    &\multicolumn{5}{c}{AIGC Semantic Edit}\\
\cmidrule(lr){3-7}	\cmidrule(lr){8-12}
&&GN &JPEG   &Contr   &Bright &GF   &LaMa   &RD   &SDIP &P2P    &SDIP+\\
\midrule
noVAE & 38.91/0.990 &97.20&	91.77&	99.47&	99.62&	99.77&	86.78&	87.42&	86.06&	82.19&	77.06\\
\textbf{with VAE} & \textbf{39.19/0.990} 	&\textbf{99.20}&	\textbf{94.31}&	\textbf{99.94}&	\textbf{99.84}&	\textbf{99.86}&	\textbf{94.22}&	\textbf{93.86}&	\textbf{94.02}&	\textbf{90.42}&	\textbf{92.25}\\
\bottomrule
\end{tabular}
\end{table*}

\subsubsection{Visual Quality Comparison}
To demonstrate the superiority of SimuFreeMark in image quality, Fig.~\ref{fig:visual} presents a visual comparison between the watermarked images and the residual images (magnified by 10×). The baseline methods introduce varying degrees of visual artifacts: HiDDeN and EditGuard show color shifts, while PIMoG and FIN suffer from blurring, RoSteALS and Robust-Wide exhibit localized noise patterns.

In contrast, the watermarked image of SimuFreeMark is visually indistinguishable from the original. This fidelity is further confirmed by its residual images, which reveals minimal and structurally coherent perturbations. These results demonstrate that by embedding watermarks into the structurally stable low-frequency components, our method effectively avoids introducing perceptible distortions and achieves superior visual quality.

\begin{table*}[!t]
\centering
\small
\caption{Ablation study on noise simulation.}
\label{tab:comparison_noiselayer}
    \vspace{-3pt}
\begin{tabular}{lccccccccccc}
\toprule
\multirow{2}{*}{Method}	&\multirow{2}{*}{PSNR/SSIM} &\multicolumn{5}{c}{Conventional Attack}    &\multicolumn{5}{c}{AIGC Semantic Edit}\\
\cmidrule(lr){3-7}	\cmidrule(lr){8-12}
&&GN &JPEG   &Contr   &Bright &GF   &LaMa   &RD   &SDIP &P2P    &SDIP+\\
\midrule
Simu & 37.93/0.984 	&98.42&	89.09&	98.78&	99.63&	99.80&	93.33&	\textbf{94.80}&	92.48&	\textbf{94.39}&	\textbf{92.50}\\
\textbf{SimuFree} & \textbf{39.19/0.990} 	&\textbf{99.20}&	\textbf{94.31}&	\textbf{99.94}&	\textbf{99.84}&	\textbf{99.86}&	\textbf{94.22}&	93.86&	\textbf{94.02}&	90.42&	92.25\\
\bottomrule
\end{tabular}
\end{table*}

\subsection{Ablation Study}
\label{subsec:abla}
To validate the design in the proposed framework, we conduct a series of ablation studies. The experiments investigate the impact of different low-frequency component extractor, the role of the pre-trained VAE, and the effect of introducing a noise simulation layer during training.

For all ablation studies, the attack parameters are fixed as: GN ($\sigma$=0.2), JPEG ($q$=10), Contrast ($r$=$\pm$ 40\%), Brightness ($r$=$\pm$ 30\%), GF ($k$=7), SDIP ($s$=0.7), SDIP+ ($s$=0.5). LaMa, RD, and P2P are performed with the default settings.

\subsubsection{Impact of Low-Frequency Component Extractor}
We evaluate three methods to isolate the low-frequency component: DCT, which reconstructs the image from the lowest 15\% of DCT coefficients; GAU, which applies Gaussian filtering (15$\times$15 kernel, $\sigma$=2.5) to obtain a low-pass version; and our FFT-based approach.

The results in Table~\ref{tab:comparison_extra_method} demonstrate that all three extractors achieve high fidelity (PSNR $>$38.9 dB, SSIM $>$0.988) and robust performance, confirming that any stable low-frequency component can serve as an effective foundation. 

Among them, the FFT-based extractor is superior, particularly against AIGC edits. For example, it outperforms both DCT and GAU extractors on LaMa. We attribute this to the global, phase-aware nature of the FFT, which preserves structural integrity more effectively than block-wise DCT or the spatial blurring induced by Gaussian filtering. This capability to maintain global structure is crucial for watermark survival under semantic manipulations that alter local content while preserving the overall scene composition.

\subsubsection{Impact of the Pre-Trained VAE}
We evaluate the importance of the pre-trained VAE by comparing our full model against an ablated version (noVAE). In noVAE, the watermark is embedded/extracted directly in the spatial domain of the low-frequency image, replacing the VAE's latent space with other components unchanged.

The results in Table~\ref{tab:comparison_vae} show that both variants achieve high visual quality. However, the model with VAE exhibits significantly stronger robustness, especially against AIGC edits. For example, it outperforms the noVAE version on LaMa (94.22\% vs. 86.78\%) and SDIP+ (92.25\% vs. 77.06\%). This demonstrates that embedding in the VAE's deep feature space, which captures semantic structural, is crucial for resisting semantic manipulations. In contrast, operating directly on pixels lacks this semantic association, making the watermark more vulnerable to AIGC edits.

\subsubsection{Impact of the Noise Simulation Layer}
We validate the effectiveness of our simulation-free algorithm (SimuFree) by comparing it with a variant (Simu) that was trained using a composite noise layer that included GN ($\sigma$=0.2), JPEG ($q$=10), GF ($k$=7), RD, and P2P attacks.

The results in Table~\ref{tab:comparison_noiselayer} demonstrate that incorporating the noise layer degrades visual quality (PSNR: 37.93 dB vs. 39.19 dB). As anticipated, the Simu variant achieves superior performance on the specific AIGC edits it was trained on (e.g., RD and P2P), confirming that targeted simulation is effective for known attacks. In contrast, our method attains higher fidelity and more stable performance across both conventional signal processing and diverse AIGC edits. This illustrates a key trade-off: while targeted simulation can optimize for specific, known threats, our simulation-free approach, by leveraging inherent low-frequency stability, provides broader and more generalizable robustness without compromising visual quality.

\section{Conclusion}
\label{sec:con}
In this paper, we propose SimuFreeMark, a novel watermarking framework that eliminates the reliance on hand-crafted noise simulation. By exploiting the inherent stability of image low-frequency components and embedding watermarks in their deep feature space via a pre-trained VAE, our method achieves robustness without being trained on any specific attack simulations.

Extensive experiments demonstrate that SimuFreeMark achieves high visual quality and effective robustness against both conventional signal processing and AIGC semantic edits. The results establish the simulation-free paradigm as a generalizable alternative to methods that depend on modeling specific attack distributions. Future work will extend this paradigm to handle geometric transformations by exploring geometry-invariant feature representations, further enhancing its practicality in real-world scenarios.

{\small
\bibliographystyle{ieee_fullname}
\bibliography{CVPR2026}

@inproceedings{alamSpecGuardSpectralProjectionbased2025,
  title = {{{SpecGuard}}: {{Spectral}} Projection-Based Advanced Invisible Watermarking},
  shorttitle = {{{SpecGuard}}},
  booktitle = {Proceedings of the {{IEEE}}/{{CVF International Conference}} on {{Computer Vision}}},
  author = {Alam, Inzamamul and Islam, Md Tanvir and Woo, Simon S. and Muhammad, Khan},
  year = 2025,
  pages = {17984--17993},
  urldate = {2025-11-12},
  langid = {english}
}

@inproceedings{brooksInstructPix2PixLearningFollow2023,
  title = {{{InstructPix2Pix}}: {{Learning}} to Follow Image Editing Instructions},
  shorttitle = {{{InstructPix2Pix}}},
  booktitle = {Proceedings of the {{IEEE}}/{{CVF Conference}} on {{Computer Vision}} and {{Pattern Recognition}}},
  author = {Brooks, Tim and Holynski, Aleksander and Efros, Alexei A.},
  year = 2023,
  pages = {18392--18402},
  urldate = {2025-09-11},
  langid = {english}
}

@inproceedings{buiRoSteALSRobustSteganography2023,
  title = {{{RoSteALS}}: {{Robust}} Steganography Using Autoencoder Latent Space},
  shorttitle = {{{RoSteALS}}},
  booktitle = {Proceedings of the {{IEEE}}/{{CVF Conference}} on {{Computer Vision}} and {{Pattern Recognition}}},
  author = {Bui, Tu and Agarwal, Shruti and Yu, Ning and Collomosse, John},
  year = 2023,
  pages = {933--942},
  urldate = {2025-11-13},
  langid = {english}
}

@inproceedings{fangFlowbasedRobustWatermarking2023,
  title = {Flow-Based Robust Watermarking with Invertible Noise Layer for Black-Box Distortions},
  booktitle = {Proceedings of the {{AAAI Conference}} on {{Artificial Intelligence}}},
  author = {Fang, Han and Qiu, Yupeng and Chen, Kejiang and Zhang, Jiyi and Zhang, Weiming and Chang, Ee-Chien},
  year = 2023,
  month = jun,
  volume = {37},
  pages = {5054--5061},
  doi = {10.1609/aaai.v37i4.25633},
  urldate = {2025-01-16},
  copyright = {Copyright (c) 2023 Association for the Advancement of Artificial Intelligence},
  langid = {english}
}

@inproceedings{fangPIMoGEffectiveScreenshooting2022,
  title = {{{PIMoG}}: {{An}} Effective Screen-Shooting Noise-Layer Simulation for Deep-Learning-Based Watermarking Network},
  shorttitle = {{{PIMoG}}},
  booktitle = {Proceedings of the 30th {{ACM International Conference}} on {{Multimedia}}},
  author = {Fang, Han and Jia, Zhaoyang and Ma, Zehua and Chang, Ee-Chien and Zhang, Weiming},
  year = 2022,
  month = oct,
  pages = {2267--2275},
  doi = {10.1145/3503161.3548049},
  urldate = {2025-02-24},
  isbn = {978-1-4503-9203-7},
  langid = {english}
}

@misc{ganGenPTWIngenerationImage2025,
  title = {{{GenPTW}}: {{In-generation}} Image Watermarking for Provenance Tracing and Tamper Localization},
  shorttitle = {{{GenPTW}}},
  author = {Gan, Zhenliang and Liu, Chunya and Tang, Yichao and Wang, Binghao and Wang, Weiqiang and Zhang, Xinpeng},
  year = 2025,
  month = apr,
  number = {arXiv:2504.19567},
  eprint = {2504.19567},
  primaryclass = {cs},
  publisher = {arXiv},
  doi = {10.48550/arXiv.2504.19567},
  urldate = {2025-09-11},
  archiveprefix = {arXiv},
  langid = {english}
}

@article{huoDRSWDualstageRobust2025,
  title = {{{DRSW}}: {{Dual-stage}} Robust Semantic Watermarking for Image Semantic Communication},
  shorttitle = {{{DRSW}}},
  author = {Huo, Yanhao and Xiang, Shijun},
  year = 2025,
  journal = {IEEE Transactions on Circuits and Systems for Video Technology},
  pages = {1--1},
  issn = {1051-8215},
  doi = {10.1109/TCSVT.2025.3613856},
  urldate = {2025-11-12},
  langid = {english},
  lccn = {1 (CCF B)}
}

@inproceedings{huRobustwideRobustWatermarking2025,
  title = {Robust-Wide: Robust Watermarking against Instruction-Driven Image Editing},
  shorttitle = {Robust-Wide},
  booktitle = {European Conference on Computer Vision},
  author = {Hu, Runyi and Zhang, Jie and Xu, Ting and Li, Jiwei and Zhang, Tianwei},
  year = 2025,
  pages = {20--37},
  doi = {doi.org/10.1007/978-3-031-72670-5_2},
  isbn = {978-3-031-72670-5},
  langid = {english}
}

@inproceedings{jiaMBRSEnhancingRobustness2021,
  title = {{{MBRS}}: {{Enhancing}} Robustness of {{DNN-based}} Watermarking by Mini-Batch of Real and Simulated {{JPEG}} Compression},
  shorttitle = {{{MBRS}}},
  booktitle = {Proceedings of the 29th {{ACM International Conference}} on {{Multimedia}}},
  author = {Jia, Zhaoyang and Fang, Han and Zhang, Weiming},
  year = 2021,
  month = oct,
  pages = {41--49},
  doi = {10.1145/3474085.3475324},
  urldate = {2025-01-16},
  isbn = {978-1-4503-8651-7},
  langid = {english}
}

@inproceedings{klambauerSelfnormalizingNeuralNetworks2017,
  title = {Self-Normalizing Neural Networks},
  booktitle = {Advances in {{Neural Information Processing Systems}}},
  author = {Klambauer, G{\"u}nter and Unterthiner, Thomas and Mayr, Andreas and Hochreiter, Sepp},
  year = 2017,
  volume = {30},
  publisher = {Curran Associates, Inc.},
  urldate = {2025-11-13},
  langid = {english}
}

@inproceedings{linMicrosoftCOCOCommon2014,
  title = {Microsoft {{COCO}}: {{Common}} Objects in Context},
  shorttitle = {Microsoft {{COCO}}},
  booktitle = {European Conference on Computer Vision},
  author = {Lin, Tsung-Yi and Maire, Michael and Belongie, Serge and Hays, James and Perona, Pietro and Ramanan, Deva and Doll{\'a}r, Piotr and Zitnick, C. Lawrence},
  year = 2014,
  pages = {740--755},
  doi = {10.1007/978-3-319-10602-1_48},
  isbn = {978-3-319-10602-1},
  langid = {english}
}

@inproceedings{luoIRWArtLeveringWatermarking2023,
  title = {{{IRWArt}}: Levering Watermarking Performance for Protecting High-Quality Artwork Images},
  shorttitle = {{{IRWArt}}},
  booktitle = {Proceedings of the {{ACM}} Web Conference 2023},
  author = {Luo, Yuanjing and Zhou, Tongqing and Liu, Fang and Cai, Zhiping},
  year = 2023,
  month = apr,
  pages = {2340--2348},
  doi = {10.1145/3543507.3583489},
  urldate = {2025-01-16},
  isbn = {978-1-4503-9416-1},
  langid = {english}
}

@inproceedings{luRobustWatermarkingUsing2024,
  title = {Robust Watermarking Using Generative Priors against Image Editing: {{From}} Benchmarking to Advances},
  shorttitle = {Robust Watermarking Using Generative Priors against Image Editing},
  booktitle = {The {{Thirteenth International Conference}} on {{Learning Representations}}},
  author = {Lu, Shilin and Zhou, Zihan and Lu, Jiayou and Zhu, Yuanzhi and Kong, Adams Wai-Kin},
  year = 2024,
  month = oct,
  urldate = {2025-09-11},
  langid = {english}
}

@inproceedings{maBlindWatermarkingCombining2022,
  title = {Towards Blind Watermarking: {{Combining}} Invertible and Non-Invertible Mechanisms},
  shorttitle = {Towards Blind Watermarking},
  booktitle = {Proceedings of the 30th {{ACM}} International Conference on Multimedia},
  author = {Ma, Rui and Guo, Mengxi and Hou, Yi and Yang, Fan and Li, Yuan and Jia, Huizhu and Xie, Xiaodong},
  year = 2022,
  month = oct,
  pages = {1532--1542},
  doi = {10.1145/3503161.3547950},
  urldate = {2025-01-16},
  isbn = {978-1-4503-9203-7},
  langid = {english}
}

@inproceedings{maRoPaSSRobustWatermarking2025,
  title = {{{RoPaSS}}: {{Robust}} Watermarking for Partial Screen-Shooting Scenarios},
  booktitle = {Proceedings of the {{AAAI Conference}} on {{Artificial Intelligence}}},
  author = {Ma, Zehua and Fang, Han and Yang, Xi and Chen, Kejiang and Zhang, Weiming},
  year = 2025,
  volume = {39},
  pages = {19332--19339},
  doi = {10.1609/aaai.v39i18.34128},
  langid = {english}
}

@article{qinPrintcameraResistantImage2024,
  title = {Print-Camera Resistant Image Watermarking with Deep Noise Simulation and Constrained Learning},
  author = {Qin, Chuan and Li, Xiaomeng and Zhang, Zhenyi and Li, Fengyong and Zhang, Xinpeng and Feng, Guorui},
  year = 2024,
  journal = {IEEE Transactions on Multimedia},
  volume = {26},
  pages = {2164--2177},
  issn = {1520-9210},
  doi = {10.1109/TMM.2023.3293272},
  urldate = {2024-02-19},
  langid = {english},
  lccn = {1}
}

@inproceedings{raoDynMarkRobustWatermarking2025,
  title = {{{DynMark}}: {{A}} Robust Watermarking Solution for Dynamic Screen Content with Small-Size Screenshot Support},
  shorttitle = {{{DynMark}}},
  booktitle = {Proceedings of the 33rd {{ACM International Conference}} on {{Multimedia}}},
  author = {Rao, Changyu and Liu, Gaozhi and Li, Sheng and Zhang, Xinpeng and Qian, Zhenxing},
  year = 2025,
  month = oct,
  pages = {7463--7471},
  doi = {10.1145/3746027.3754897},
  urldate = {2025-11-11},
  isbn = {979-8-4007-2035-2},
  langid = {english}
}

@inproceedings{rombachHighresolutionImageSynthesis2022,
  title = {High-Resolution Image Synthesis with Latent Diffusion Models},
  booktitle = {Proceedings of the {{IEEE}}/{{CVF Conference}} on {{Computer Vision}} and {{Pattern Recognition}}},
  author = {Rombach, Robin and Blattmann, Andreas and Lorenz, Dominik and Esser, Patrick and Ommer, Bj{\"o}rn},
  year = 2022,
  pages = {10684--10695},
  urldate = {2025-09-11},
  langid = {english}
}

@article{shijunxiangInvariantImageWatermarking2008,
  title = {Invariant Image Watermarking Based on Statistical Features in the Low-Frequency Domain},
  author = {Shijun Xiang and Hyoung Joong Kim and Jiwu Huang},
  year = 2008,
  month = jun,
  journal = {IEEE Transactions on Circuits and Systems for Video Technology},
  volume = {18},
  number = {6},
  pages = {777--790},
  issn = {1051-8215},
  doi = {10.1109/TCSVT.2008.918843},
  urldate = {2025-11-12},
  langid = {english},
  lccn = {1 (CCF B)}
}

@inproceedings{suvorovResolutionrobustLargeMask2022,
  title = {Resolution-Robust Large Mask Inpainting with Fourier Convolutions},
  booktitle = {Proceedings of the {{IEEE}}/{{CVF Winter Conference}} on {{Applications}} of {{Computer Vision}}},
  author = {Suvorov, Roman and Logacheva, Elizaveta and Mashikhin, Anton and Remizova, Anastasia and Ashukha, Arsenii and Silvestrov, Aleksei and Kong, Naejin and Goka, Harshith and Park, Kiwoong and Lempitsky, Victor},
  year = 2022,
  pages = {2149--2159},
  urldate = {2025-09-11},
  langid = {english}
}

@article{tangNovelRobustReversible2025,
  title = {A Novel Robust Reversible Watermarking Scheme Using Fractional-Order Polar Complex Exponential Transform},
  author = {Tang, Yichao and Wang, Shuai and Han, Ruowei and Wang, Chuntao},
  year = 2025,
  journal = {IEEE Transactions on Multimedia},
  pages = {1--15},
  issn = {1520-9210},
  doi = {10.1109/TMM.2025.3623522},
  urldate = {2025-11-12},
  langid = {english},
  lccn = {1 (CCF B)}
}

@article{tangRobustReversibleWatermarking2024,
  title = {A Robust Reversible Watermarking Scheme Using Attack-Simulation-Based Adaptive Normalization and Embedding},
  author = {Tang, Yichao and Wang, Chuntao and Xiang, Shijun and Cheung, Yiu-Ming},
  year = 2024,
  journal = {IEEE Transactions on Information Forensics and Security},
  volume = {19},
  pages = {4114--4129},
  issn = {1556-6013},
  doi = {10.1109/TIFS.2024.3372811},
  urldate = {2024-04-20},
  langid = {english},
  lccn = {1}
}

@article{wuEnhancedJustNoticeable2017,
  title = {Enhanced Just Noticeable Difference Model for Images with Pattern Complexity},
  author = {Wu, Jinjian and Li, Leida and Dong, Weisheng and Shi, Guangming and Lin, Weisi and Kuo, C.-C. Jay},
  year = 2017,
  month = jun,
  journal = {IEEE Transactions on Image Processing},
  volume = {26},
  number = {6},
  pages = {2682--2693},
  issn = {1057-7149},
  doi = {10.1109/TIP.2017.2685682},
  urldate = {2025-11-13},
  langid = {english},
  lccn = {1 (CCF A)}
}

@inproceedings{zhangArtBankArtisticStyle2024,
  title = {{{ArtBank}}: {{Artistic}} Style Transfer with Pre-Trained Diffusion Model and Implicit Style Prompt Bank},
  shorttitle = {{{ArtBank}}},
  booktitle = {Proceedings of the {{AAAI Conference}} on {{Artificial Intelligence}}},
  author = {Zhang, Zhanjie and Zhang, Quanwei and Xing, Wei and Li, Guangyuan and Zhao, Lei and Sun, Jiakai and Lan, Zehua and Luan, Junsheng and Huang, Yiling and Lin, Huaizhong},
  year = 2024,
  month = mar,
  volume = {38},
  pages = {7396--7404},
  doi = {10.1609/aaai.v38i7.28570},
  urldate = {2025-02-25},
  copyright = {Copyright (c) 2024 Association for the Advancement of Artificial Intelligence},
  langid = {english}
}

@inproceedings{zhangEditGuardVersatileImage2024,
  title = {{{EditGuard}}: {{Versatile}} Image Watermarking for Tamper Localization and Copyright Protection},
  shorttitle = {{{EditGuard}}},
  booktitle = {Proceedings of the {{IEEE}}/{{CVF}} Conference on Computer Vision and Pattern Recognition},
  author = {Zhang, Xuanyu and Li, Runyi and Yu, Jiwen and Xu, Youmin and Li, Weiqi and Zhang, Jian},
  year = 2024,
  pages = {11964--11974},
  urldate = {2025-01-16},
  langid = {english}
}

@inproceedings{zhaoUltraEditInstructionbasedFinegrained2024,
  title = {{{UltraEdit}}: {{Instruction-based}} Fine-Grained Image Editing at Scale},
  shorttitle = {{{UltraEdit}}},
  booktitle = {The {{Thirty-eight Conference}} on {{Neural Information Processing Systems Datasets}} and {{Benchmarks Track}}},
  author = {Zhao, Haozhe and Ma, Xiaojian and Chen, Liang and Si, Shuzheng and Wu, Rujie and An, Kaikai and Yu, Peiyu and Zhang, Minjia and Li, Qing and Chang, Baobao},
  year = 2024,
  month = nov,
  urldate = {2025-09-10},
  langid = {english}
}

@inproceedings{zhuHiDDeNHidingData2018,
  title = {{{HiDDeN}}: {{Hiding}} Data with Deep Networks},
  shorttitle = {{{HiDDeN}}},
  booktitle = {Proceedings of the European Conference on Computer Vision},
  author = {Zhu, Jiren and Kaplan, Russell and Johnson, Justin and Li, Fei-Fei},
  year = 2018,
  pages = {657--672},
  urldate = {2025-02-25},
  langid = {english}
}

@misc{wuSimtorealUnsupervisedNoise2025,
  title = {Sim-to-Real: {{An}} Unsupervised Noise Layer for Screen-Camera Watermarking Robustness},
  shorttitle = {Sim-to-Real},
  author = {Wu, Yufeng and Liao, Xin and Wang, Baowei and Fang, Han and Wu, Xiaoshuai and Chen, Mingyue and Wang, Guiling},
  year = 2025,
  month = apr,
  number = {arXiv:2504.18906},
  eprint = {2504.18906},
  primaryclass = {cs},
  publisher = {arXiv},
  doi = {10.48550/arXiv.2504.18906},
  urldate = {2025-06-14},
  archiveprefix = {arXiv},
  langid = {english}
}

@inproceedings{fangDERODiffusionmodelerasureRobust2024,
  title = {{{DERO}}: {{Diffusion-model-erasure}} Robust Watermarking},
  shorttitle = {{{DERO}}},
  booktitle = {Proceedings of the 32nd {{ACM}} International Conference on Multimedia},
  author = {Fang, Han and Chen, Kejiang and Qiu, Yupeng and Ma, Zehua and Zhang, Weiming and Chang, Ee-Chien},
  year = 2024,
  month = oct,
  pages = {2973--2981},
  doi = {10.1145/3664647.3681220},
  urldate = {2025-02-24},
  isbn = {979-8-4007-0686-8},
  langid = {english}
}

@article{gaoEfficientRobustReversible2024,
  title = {Efficient Robust Reversible Watermarking Based on {{ZMs}} and Integer Wavelet Transform},
  author = {Gao, Guangyong and Wang, Min and Wu, Bin},
  year = 2024,
  journal = {IEEE Transactions on Industrial Informatics},
  pages = {1--9},
  issn = {1551-3203},
  doi = {10.1109/TII.2023.3321101},
  urldate = {2024-01-27},
  langid = {english},
  lccn = {10.215}
}

@String(AAAI = {AAAI})
}

\end{document}